\documentclass{article}

\usepackage{iclr2027_conference,times}

\usepackage{amsmath,amsfonts,bm}

\def\eqref#1{equation~\ref{#1}}

\def\1{\bm{1}}

\DeclareMathAlphabet{\mathsfit}{\encodingdefault}{\sfdefault}{m}{sl}
\SetMathAlphabet{\mathsfit}{bold}{\encodingdefault}{\sfdefault}{bx}{n}

\usepackage{amsmath}
\usepackage{booktabs}
\usepackage{array}
\usepackage{colortbl}
\usepackage{flafter}
\usepackage{float}
\usepackage{dblfloatfix}
\usepackage{graphicx}
\usepackage{xcolor}
\usepackage{microtype}
\usepackage{hyperref}
\usepackage{url}
\hypersetup{hidelinks}
\usepackage{wrapfig}

\definecolor{gaincolor}{RGB}{0,125,75}
\definecolor{dropcolor}{RGB}{190,45,45}
\definecolor{baselinebg}{RGB}{247,247,247}
\definecolor{oursbg}{RGB}{239,246,252}
\newcommand{\gain}[1]{\textcolor{gaincolor}{\scriptsize\,(+#1)}}
\newcommand{\dropresult}[1]{\textcolor{dropcolor}{\scriptsize\,(-#1)}}

\title{PREM: Prediction-Regulated Memory for Robot Manipulation}

\author{%
  Xiao Huang \quad Ya Zhang\textsuperscript{\dag} \quad Weidi Xie\textsuperscript{\dag}\\
  \normalfont School of Artificial Intelligence, Shanghai Jiao Tong University\\
  \normalfont\small \textsuperscript{\dag}Corresponding authors
}
\iclrfinalcopy

\hypersetup{
  pdftitle={PREM: Prediction-Regulated Memory for Robot Manipulation},
  pdfauthor={Xiao Huang, Ya Zhang, Weidi Xie}
}

\renewcommand{\eqref}[1]{\textup{(\ref{#1})}}

\begin{document}

\maketitle
\thispagestyle{plain}

\begin{abstract}
A robot's next action can depend on recent motion or earlier visual
evidence absent from its current observation. A recurrent visual-memory
architecture can retain this history in a visual state, but its
usefulness depends on what the state represents and how it is updated
with new observations. We introduce \emph{Prediction-Regulated Memory
(PREM)}, a framework that uses action-conditioned future prediction
to guide both memory learning and updating in vision--language--action
policies. During training, predicting future visual features provides
temporal supervision for memory learning. During execution, PREM
compares current visual features with the prediction made at the
preceding decision and uses the spatially normalized prediction
discrepancy to regulate updates to the visual state at each spatial
location. The updated visual state supplies temporal context for
action generation alongside current observations. Built on
$\pi_{0.5}$, PREM improves average success rate from 9.63\% to 20.74\%
on DOMINO dynamic manipulation tasks and from 14.4\% to 83.8\% on the
RMBench M(1) memory-dependent setting. Further evaluations on LIBERO
and LIBERO-Pro show improvements in general manipulation and robustness
to spatial perturbations. PREM incurs a mean inference overhead of
7.7\,ms on a single NVIDIA H800 GPU.
\href{https://1106280506hx.github.io/PREM/}{\textbf{Project page}}.
\end{abstract}

\newcommand{\weidi}[1]{{\textcolor{magenta}{[WEIDI: #1]}}}

\section{Introduction}
\label{sec:introduction}

\begin{figure}[!t]
    \centering
    \includegraphics[width=\linewidth]
    {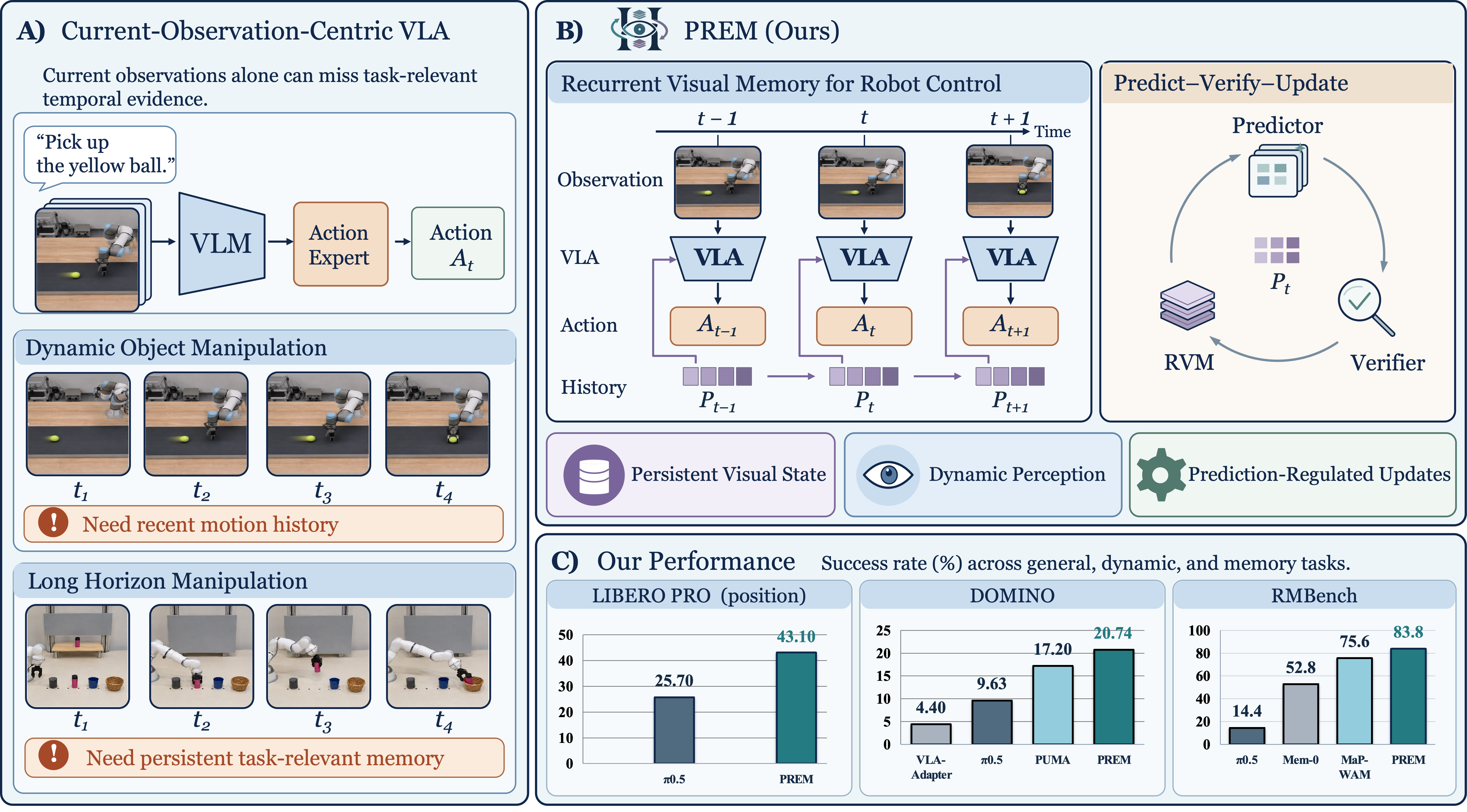}
    \caption{\textbf{Motivation and overview of PREM.}
    \textbf{(A)} Current observations alone can miss recent motion
    and earlier task-relevant visual evidence.
    \textbf{(B)} PREM uses future prediction to supervise memory
    learning and prediction discrepancy to regulate visual-state
    updating.
    \textbf{(C)} PREM improves general manipulation, dynamic
    manipulation, and memory-dependent manipulation.}
    \vspace{-4pt}
    \label{fig:teaser}
\end{figure}

Robot manipulation often takes place under partial observability, where the current observation may not contain all information needed for action selection. Visual observations, language instructions, and proprioceptive readings can be insufficient when relevant motion or task state must be inferred from earlier interaction. The appropriate action can therefore depend on what the robot observed and did previously. This motivates integrating current observations with task-relevant interaction history.

For example, intercepting a moving object requires information about
its recent motion, even when the object's current position is visible.
Similarly, restoring an earlier arrangement requires remembering where
objects were previously located, even when the current scene looks
similar to another state. In both cases, similar current observations
may require different actions because they correspond to different
visual histories. A manipulation policy therefore needs to combine
current observations with task-relevant visual history. \autoref{fig:teaser}
illustrates these complementary requirements.

Existing approaches provide access to visual history in different
ways. Multi-frame models process a window of recent observations
\citep{li2023roboflamingo,li2025cronusvla}, while memory-based
policies store or summarize earlier evidence
\citep{fang2025sam2act,shi2025memoryvla,huang2026chainvla}. We focus
on a recurrent visual-memory architecture that carries a fixed-size
visual state across action decisions. However, retaining history alone
is not sufficient. The visual state must preserve information relevant
to future action generation, while incorporating new observations
without discarding useful history. These requirements concern both
what the visual state represents and how it is updated during
interaction.

Future prediction has been used to learn visual representations through
self-supervised objectives~\citep{han2019video,han2020memory}.
Recent VLA methods combine future prediction with action generation
and memory~\citep{sun2026vlajepa,shi2026memoryvlapp,yuan2026fastwam}.
Related video-understanding work also studies recognizing, localizing,
and anticipating unintentional actions~\citep{epstein2020oops}. Future
prediction provides a natural way to address both requirements.
During training, predicting subsequent visual features encourages the
visual state to retain information useful for anticipating interaction
dynamics. The prediction targets come directly from future
observations, providing temporal supervision without additional
annotations. During execution, the discrepancy between predicted and
observed visual features can reveal where the current visual state is
insufficient. This motivates using action-conditioned future
prediction not only to supervise memory learning, but also to guide
visual-state updating.

This leads to our central question:
\emph{how can action-conditioned future prediction guide both memory
learning and visual-state updating for robot manipulation?}

We introduce \emph{Prediction-Regulated Memory (PREM)}, a framework
that uses action-conditioned future prediction to guide both memory
learning and visual-state updating in VLA policies. PREM couples a
recurrent visual-memory (RVM) architecture with an
action-conditioned predictor. The RVM architecture maintains a
fixed-size visual state across action decisions. The policy, RVM
architecture, and predictor are jointly trained with action and
prediction objectives. The prediction objective provides temporal
supervision for memory learning, while the discrepancy between
prediction and observation provides feedback for visual-state
updating.

At each replanning decision, PREM compares the current visual
features with the prediction made at the preceding decision. The
spatially normalized prediction discrepancy regulates how much of the
candidate visual-state update is incorporated at each spatial
location. The updated visual state is then used as temporal context
for action generation together with the current observation. The
policy's latent action representation is also provided to the
predictor, which predicts the visual features of the next replanning
observation.

During training, the prediction objective supervises memory learning
through the predictor's visual-state input. During execution, the
prediction is cached until the corresponding observation arrives and
then provides the reference for the next visual-state update. In this
way, PREM uses the same future-prediction mechanism to learn the
visual state and to regulate how it changes over time.

Our evaluation follows the two temporal requirements in
\autoref{fig:teaser}. We test dynamic manipulation on DOMINO and
memory-dependent manipulation on RMBench M(1). Built on $\pi_{0.5}$,
PREM improves average success rate from 9.63\% to 20.74\% on DOMINO
and from 14.4\% to 83.8\% on RMBench M(1). Component ablations on
DOMINO show performance gains from both predictive supervision for
memory learning and prediction-regulated visual-state updating.
Additional evaluations on LIBERO and LIBERO-Pro show improvements
in general manipulation and robustness to spatial perturbations.
PREM incurs a mean inference overhead of 7.7\,ms on a single
NVIDIA H800 GPU.

\section{Method}
\label{sec:method}

\begin{figure}[t]
    \centering
    \includegraphics[width=\linewidth]
    {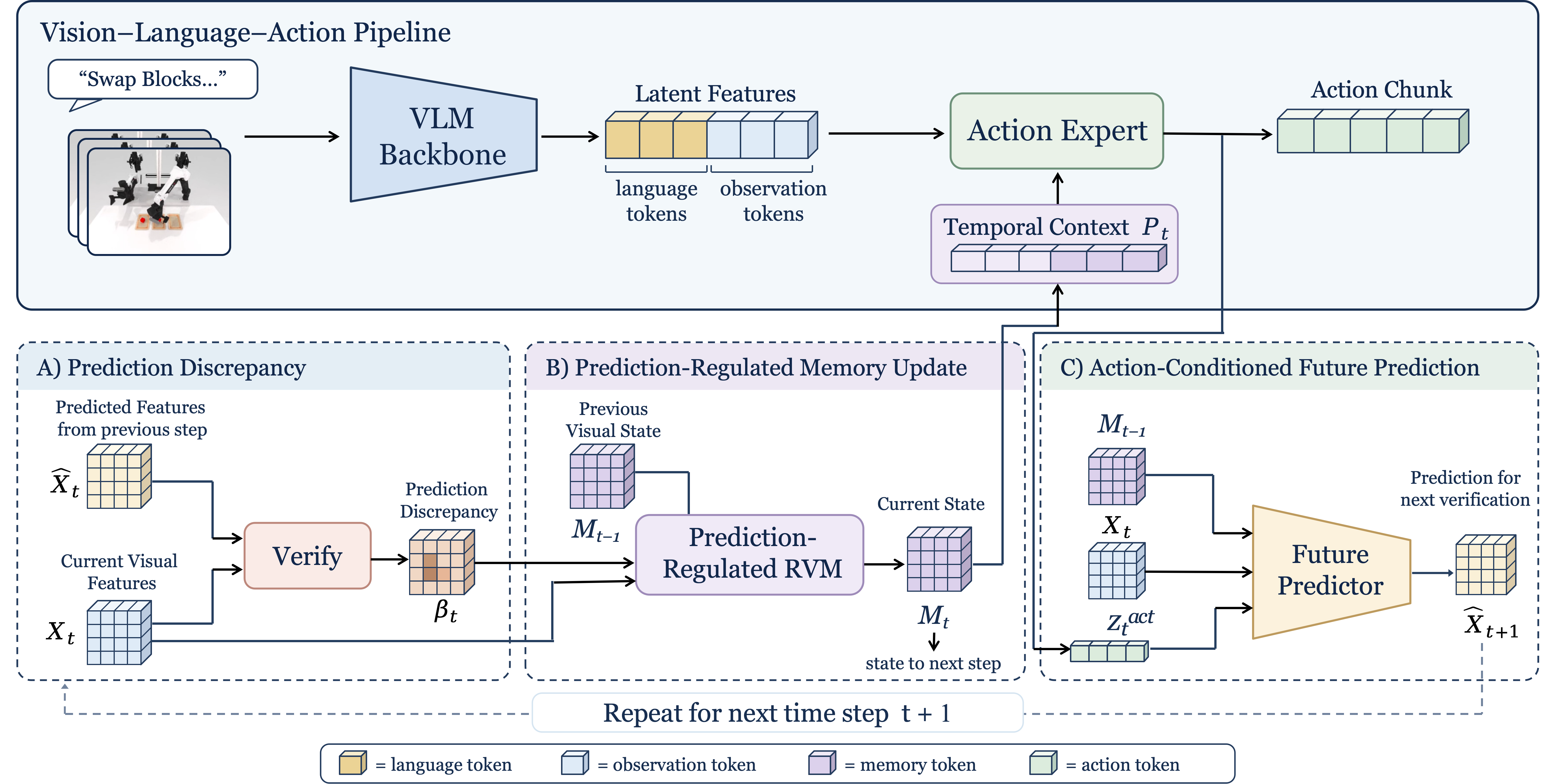}
    \caption{\textbf{PREM architecture.}
    \textbf{(A)} PREM compares the cached prediction with observed
    visual features at each spatial location to compute prediction
    discrepancy.
    \textbf{(B)} The spatially normalized discrepancy regulates how
    much of the candidate visual-state update is incorporated.
    \textbf{(C)} An action-conditioned predictor predicts the next
    visual features from the incoming visual state, observed visual
    features, and latent action representation. The prediction
    objective supervises memory learning. The cached prediction
    provides a reference for subsequent visual-state updating.}
    \label{fig:method-architecture}
\end{figure}

PREM couples an action-conditioned predictor with a recurrent
visual-memory (RVM) architecture. Future prediction serves two
complementary roles: its training objective supervises memory learning,
while its discrepancy from subsequent observations provides feedback
for visual-state updating. At each replanning decision, PREM receives
the current observation, verifies the prediction cached at the
preceding decision, updates the visual state, generates actions, and
predicts the next visual observation. \autoref{fig:method-architecture}
summarizes this prediction--verification--update cycle.

\subsection{Problem Formulation}
\label{sec:temporal_state}

We consider language-conditioned robot manipulation under partial
observability. At each replanning decision $t$, the robot receives
visual observations $O_t$, robot proprioception $s_t$, and a language
instruction $\ell$. The current observation may not contain recent
motion or earlier task-relevant visual evidence. The base VLA policy
predicts an action chunk $A_t$, a sequence of robot control commands:
\begin{equation}
A_t \sim \pi_{\mathrm{base}}(\cdot\mid O_t,s_t,\ell).
\label{eq:base_policy}
\end{equation}
The robot executes a selected prefix of $A_t$, then obtains a new
observation and replans at $t+1$. Thus, $t$ indexes replanning
decisions rather than individual control steps. PREM retains the
relevant history in a fixed-size visual state. At decision $t$, it
(1) extracts current visual features, (2) compares them with the
prediction cached at the preceding decision, (3) updates the visual
state, (4) uses the updated state for action generation, and (5)
predicts the visual features at the next decision.

Temporal context $P_t$ derived from the updated visual state augments
the original policy inputs:
\begin{equation}
A_t \sim \pi_\theta(\cdot\mid O_t,s_t,\ell,P_t).
\label{eq:action_generation}
\end{equation}
The following sections describe the prediction discrepancy, the
visual-state update, the action-conditioned predictor, and the joint
training objective.

\subsection{Prediction Discrepancy}
\label{sec:prediction_discrepancy}

Prediction discrepancy supplies spatial feedback for visual-state
updating. PREM compares each new observation with the prediction
cached at the preceding decision. The comparison operates on visual
features:
\begin{equation}
X_t = \mathcal F_{\mathrm{vis}}(O_t)
\in \mathbb{R}^{N\times d},
\label{eq:visual_features}
\end{equation}
where $\mathcal F_{\mathrm{vis}}$ comprises the policy's image encoder
and a learned projection. $N$ is the number of spatial patches, and
$d$ is the feature dimension.

Let $\hat X_t$ denote the cached prediction of $X_t$. Both have the
same spatial layout, so PREM compares them at each spatial location:
\begin{equation}
e_{t,i} = \delta(\hat X_{t,i},X_{t,i}),
\qquad i=1,\ldots,N.
\label{eq:prediction_error}
\end{equation}
The function $\delta$ combines cosine distance and a mean elementwise
Smooth-L1 penalty after token-wise feature normalization. Its complete
definition is given in Appendix~B. No external motion estimator is
required.

We use relative spatial discrepancy to regulate updates at different
locations. The spatial median is
$\mu_t=\operatorname{median}_j e_{t,j}$, and the median absolute
deviation is $\rho_t=\operatorname{median}_j|e_{t,j}-\mu_t|$.
The normalized prediction discrepancy is
\begin{equation}
\xi_{t,i}
=
\frac{e_{t,i}-\mu_t}{\max(\rho_t,\epsilon_m)},
\label{eq:normalized_surprise}
\end{equation}
where $\epsilon_m>0$ lower-bounds the denominator. Larger values
identify locations with greater discrepancy relative to the rest of
the observation. This is a spatial update signal, not an absolute
measure of frame-level unpredictability. A uniform additive shift
in the discrepancies leaves it unchanged.

\subsection{Prediction-Regulated Visual-State Update}
\label{sec:memory_update}

Visual-state updating separates the candidate content from how much
of that candidate is incorporated. The RVM architecture proposes the
candidate visual state, while prediction discrepancy regulates the
amount of the proposed change.

Let $M_{t-1}\in\mathbb{R}^{N\times d}$ denote the incoming visual
state. The RVM architecture is adapted from recurrent video
representations~\citep{zoran2025rvm}. Its gated network $R_\phi$
combines $M_{t-1}$ with $X_t$ to produce a candidate visual state
$\widetilde M_t=R_\phi(X_t,M_{t-1})$.

Visual changes poorly predicted from the available history and action
information may mark key interaction transitions. This motivates
greater incorporation of the corresponding visual evidence. PREM
implements this intuition with an update coefficient at each spatial
location:
\begin{equation}
\beta_{t,i}
=
\beta_{\min}
+
(1-\beta_{\min})
\sigma\!\left(\frac{\xi_{t,i}-\tau}{\gamma}\right),
\label{eq:update_coefficient}
\end{equation}
where $\sigma$ is the sigmoid function. The threshold $\tau$ and
transition scale $\gamma>0$ control the response to relative
discrepancy. The lower bound $0<\beta_{\min}<1$ preserves a nonzero
amount of the candidate update. The resulting visual state is
\begin{equation}
M_t
=
M_{t-1}
+
\beta_t\odot(\widetilde M_t-M_{t-1}),
\label{eq:regulated_update}
\end{equation}
where $\odot$ denotes elementwise multiplication and $\beta_t$ is
broadcast across the feature dimension.

Larger relative discrepancies incorporate more of the candidate
update; smaller ones yield more conservative updates. The coefficient
controls the amount of the proposed update, not the absolute change
in the visual state, which also depends on
$\widetilde M_t-M_{t-1}$. This modulation supplements the RVM
architecture's learned gates and requires no keyframe annotations.
At the first valid observation after a reset, we initialize
$M_0=X_0$ and skip discrepancy-based regulation because no preceding
prediction is available.

\subsection{Action-Conditioned Future Prediction}
\label{sec:future_prediction}

The predictor connects memory learning with visual-state updating.
It predicts visual features at the next replanning decision from
historical, visual, and action information. The action information
comes from the policy's own action-generation computation.

\textbf{Action conditioning.}
The updated visual state first supplies temporal context for action
generation. Let $\mathcal P_{\mathrm{mem}}$ denote the projection
used to form temporal context at the policy token width:
\begin{equation}
P_t = \mathcal P_{\mathrm{mem}}(M_t).
\label{eq:temporal_readout}
\end{equation}
The policy uses $P_t$ through Eq.~\eqref{eq:action_generation}.
Historical evidence supplements the original observation pathway
rather than replacing it. PREM reuses the final-layer action-token
features from the action expert, the policy's action-generation
module. Temporal encoding and pooling yield the latent action
representation $z_t^{\mathrm{act}}$, without another action-expert
evaluation.

\textbf{Future prediction.}
The incoming visual state $M_{t-1}$ provides historical information,
$X_t$ provides observed visual features, and $z_t^{\mathrm{act}}$
provides action information. The predictor combines these inputs:
\begin{equation}
\hat X_{t+1}
=
D_\psi(M_{t-1},X_t,z_t^{\mathrm{act}}).
\label{eq:future_prediction}
\end{equation}
Here, $D_\psi$ is a query-based Transformer decoder. Its input streams
are projected to a common width and concatenated along the token
dimension. Learned spatial queries aggregate these inputs through
cross-attention and exchange information through self-attention.
The output shares the patch layout of $X_t$. Action tokens are read
through attention, not matched one-to-one with visual patches.

The predictor directly receives the incoming state $M_{t-1}$ rather
than the updated state $M_t$. This keeps the prediction target
future-looking instead of using a state that already incorporates the
current observation. Its action input is computed after $M_t$
conditions the policy through $P_t$, so all inputs are available at
decision $t$. During training, supervision from $X_{t+1}$ propagates
through the predictor into the RVM architecture. This encourages the
visual state to retain information useful for predicting subsequent
visual features. During execution, PREM caches $\hat X_{t+1}$ until
$O_{t+1}$ arrives and then uses the prediction in the comparison of
Eq.~\eqref{eq:prediction_error}. The prediction thus supports memory
learning and subsequent update feedback; it is not supplied to the
action expert as imagined future context.

\subsection{Training and Configuration}
\label{sec:training}

We initialize from
$\pi_{0.5}$~\citep{physicalintelligence2025pi05}
and jointly fine-tune the policy, RVM architecture, and predictor.
The action objective supervises action generation. The prediction
objective supervises memory learning through the predictor's
incoming-visual-state branch.

The target $X_{t+1}$ comes from the next replanning observation via
Eq.~\eqref{eq:visual_features}. Let $v_t$ indicate a valid
prediction--target pair, and sample
$q_{t,i}\sim\operatorname{Bernoulli}(0.75)$ to select supervised
patches. The prediction loss is
\begin{equation}
\mathcal L_{\mathrm{pred}}
=
\frac{
\sum_{t,i}v_tq_{t,i}\,
\delta\!\left(
\hat X_{t+1,i},\operatorname{sg}(X_{t+1,i})
\right)
}{
\sum_{t,i}v_tq_{t,i}
},
\label{eq:prediction_loss}
\end{equation}
where $\operatorname{sg}$ denotes stop-gradient. Unlike update
regulation, this objective uses $\delta$ before spatial normalization.
The mask selects target patches for loss computation, not predictor
inputs. Targets are detached after projection; no separate or EMA
target encoder is used. The complete objective is
\begin{equation}
\mathcal L
=
\mathcal L_{\mathrm{act}}
+
\lambda_{\mathrm{pred}}\mathcal L_{\mathrm{pred}},
\label{eq:training_objective}
\end{equation}
where $\mathcal L_{\mathrm{act}}$ is the base policy's action objective
and $\lambda_{\mathrm{pred}}=0.1$.

For update regulation, $e_{t,i}$ is detached before spatial
normalization. The update coefficients therefore provide no gradient
path through the prediction discrepancy. Prediction learning remains
differentiable through the predictor's incoming visual state. Training
causally unrolls the recurrent pathway over observation clips.
Gradients propagate throughout each clip, and recurrent computation
resets between clips without a burn-in prefix.

The update parameters are fixed at $\beta_{\min}=0.25$, $\tau=2.0$,
and $\gamma=0.5$. During training, the effective coefficient starts
at one and linearly transitions to $\beta_{t,i}$ in
Eq.~\eqref{eq:update_coefficient}. Inference uses the full
prediction-regulated coefficient. Appendix~B provides architectural
specifications, action-feature extraction, discrepancy constants,
and benchmark-specific ramp, unroll, stride, and reset details.

\section{Experiments}
\raggedbottom
\label{sec:experiments}

\subsection{Experimental Setup}
\label{sec:experimental-setup}

\paragraph{Benchmarks and baselines.}
\label{sec:benchmarks}
\label{sec:baselines}
We evaluate PREM on dynamic manipulation, memory-dependent
manipulation, and general manipulation. We also assess robustness
to spatial perturbations.
LIBERO~\citep{liu2023libero} comprises four suites: LIBERO-Spatial,
LIBERO-Object, LIBERO-Goal, and LIBERO-10,covering spatial,
object-centric, goal-conditioned, and compositional manipulation.
LIBERO-Pro~\citep{zhou2025liberopro} evaluates robustness under
position-perturbed layouts and initial states. Baselines include
OpenVLA~\citep{kim2024openvla}, SpatialVLA~\citep{qu2025spatialvla},
\(\pi_0\)~\citep{black2024pi0}, \(\pi_0\)-FAST~\citep{pertsch2025fast},
GR00T-N1~\citep{nvidia2025groot}, X-VLA~\citep{zheng2025xvla},
PAM~\citep{hu2025pam}, MemoryVLA~\citep{shi2025memoryvla}, MemoryVLA++,
CronusVLA~\citep{li2025cronusvla}, and ChainVLA~\citep{huang2026chainvla}.

\begin{table}[H]
    \centering
    \caption{Success rates on standard LIBERO (\%).}
    \label{tab:libero-main}
    \small
    \setlength{\tabcolsep}{8.0pt}
    \renewcommand{\arraystretch}{1.08}
    \begin{tabular}{lccccc}
        \toprule
        Policy & Spatial & Object & Goal & LIBERO-10 & Avg. \\
        \midrule
        OpenVLA
            & 84.7 & 88.4 & 79.2 & 53.7 & 76.5 \\
        SpatialVLA
            & 88.2 & 89.9 & 78.6 & 55.5 & 78.1 \\
        \(\pi_0\)-FAST
            & 96.4 & 96.8 & 88.6 & 60.2 & 85.5 \\
        GR00T-N1
            & 94.4 & 97.6 & 93.0 & 90.6 & 93.9 \\
        \(\pi_0\)
            & 96.8 & 98.8 & 95.8 & 85.2 & 94.2 \\
        PAM
            & 86.1 & 97.0 & 94.5 & 84.7 & 90.6 \\
        MemoryVLA
            & 98.4 & 98.4 & 96.4 & 93.4 & 96.7 \\
        MemoryVLA++
            & \textbf{99.8} & \textbf{100.0}
            & 98.2 & 96.0 & 98.5 \\
        CronusVLA
            & 97.3 & \underline{99.6} & 96.9 & 94.0 & 97.0 \\
        X-VLA
            & 98.2 & 98.6 & 97.8 & \underline{97.6} & 98.1 \\
        ChainVLA
            & 98.4 & 99.4 & \underline{99.2}
            & \textbf{98.2} & \textbf{98.8} \\
        \midrule
        \rowcolor{baselinebg}
        \(\pi_{0.5}\) (Baseline)
            & 98.8 & 98.2 & 98.0 & 92.4 & 96.9 \\
        \rowcolor{oursbg}
        PREM (Ours)
            & \underline{99.2}\gain{0.4}
            & 99.0\gain{0.8}
            & \textbf{99.4}\gain{1.4}
            & 97.2\gain{4.8}
            & \underline{98.7}\gain{1.8} \\
        \bottomrule
    \end{tabular}
\end{table}

DOMINO~\citep{fang2026domino} contains 35 dynamic-interception and
dynamic-tracking tasks. Object and interaction states evolve
continuously in these tasks. We compare with \(\pi_0\),
OpenVLA-OFT~\citep{kim2025openvlaoft}, \(\pi_{0.5}\), and PUMA.
RMBench~\citep{chen2026rmbench} evaluates memory-dependent
manipulation across long temporal gaps. Baselines include
DP~\citep{chi2023diffusionpolicy}, ACT~\citep{zhao2023act}, X-VLA,
MemER~\citep{sridhar2025memer}, MemoryVLA, Mem-0, ChainVLA,
and MaP-WAM~\citep{zhao2026mapwam}. We report episode success rate.
Bold and underlined values denote the best and second-best results.
Green values indicate PREM's percentage-point gains over \(\pi_{0.5}\).

\paragraph{Training and implementation protocol.}
We initialize PREM from the official pretrained \(\pi_{0.5}\)
checkpoint. We fine-tune the complete policy end to end using
\(8\times\) NVIDIA H800 80GB GPUs. The VLM backbone and original
action-generation pathway are retained. At episode boundaries,
we reset the visual state and clear cached predictions.

Training configurations are benchmark-specific. LIBERO uses
\(1{,}693\) demonstrations, a global batch size of 128, and 30k steps.
The five RMBench tasks are trained independently. Each task uses
50 demonstrations, a global batch size of 32, and 20k steps.
DOMINO uses 1,750 episodes across 35 tasks, a global batch size
of 64, and 100k steps. For all benchmarks, the learning rate is
cosine-annealed from \(2\times10^{-5}\) to \(1\times10^{-5}\).

For latency evaluation, we measure all policies on a single
NVIDIA H800 80GB GPU with batch size 1. All policies receive the
same LIBERO input: base- and wrist-camera images, a language
instruction, and 8-dimensional robot proprioception. We perform
10 warm-up runs and report the mean over 100 consecutive runs.
Timing starts when the observation is received and ends when the
complete action chunk is available.

\subsection{Main Results}
\label{sec:main-results}

\paragraph{General manipulation.}
\label{sec:libero-evaluation}

Table~\ref{tab:libero-main} reports results on standard LIBERO.
The \(\pi_{0.5}\) base policy achieves an average success rate of
\(96.9\%\). PREM increases this to \(98.7\%\). The largest gain
occurs on LIBERO-10, from \(92.4\%\) to \(97.2\%\), an improvement
of \(4.8\) percentage points.

PREM also improves Spatial, Object, and Goal by \(0.4\), \(0.8\),
and \(1.4\) percentage points, respectively. These results show
improvements across all four suites and support the effectiveness
of PREM on general manipulation.

\paragraph{Robustness to spatial perturbations.}

\begin{wrapfigure}{r}{0.48\linewidth}
    \vspace{-12pt}
    \setlength{\columnsep}{6pt}
    \setlength{\abovecaptionskip}{2pt}
    \setlength{\belowcaptionskip}{0pt}
    \centering
    \includegraphics[width=\linewidth,trim=0 4 0 4,clip]
    {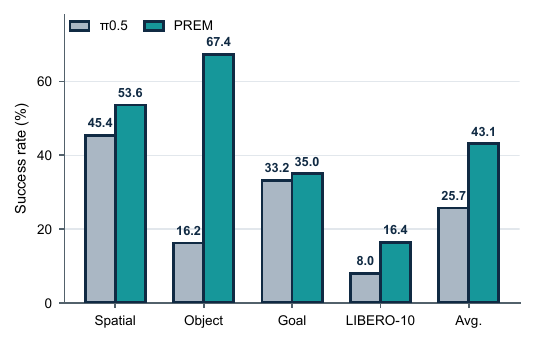}
    \caption{LIBERO-Pro success rates.}
    \label{fig:libero-pro}
    \vspace{-3pt}
\end{wrapfigure}

Figure~\ref{fig:libero-pro} reports results under LIBERO-Pro
position perturbations. PREM improves average success rate from
\(25.7\%\) to \(43.1\%\), with gains across all four suites.
The largest improvement occurs on Object, from \(16.2\%\) to
\(67.4\%\), a gain of \(51.2\) percentage points.

These results show improved robustness to position perturbations.
DOMINO and RMBench M(1) provide more targeted evaluations of
dynamic manipulation and memory-dependent manipulation,
respectively.

\begingroup
\setlength{\intextsep}{6pt}

\begin{table}[!b]
    \centering
    \caption{Representative DOMINO task success rates (\%).}
    \label{tab:domino-main}
    \footnotesize
    \setlength{\tabcolsep}{4.2pt}
    \renewcommand{\arraystretch}{1.06}
    \begin{tabular}{
        lcc
        >{\columncolor{baselinebg}}c
        c
        >{\columncolor{oursbg}}c
    }
        \toprule
        Task
        & \(\pi_0\)
        & \shortstack{OpenVLA-\\OFT}
        & \shortstack{\(\pi_{0.5}\)}
        & PUMA
        & \shortstack{PREM} \\
        \midrule

        \multicolumn{6}{l}{\emph{Dynamic Interception}} \\
        Adjust Bottle
            & 19 & 44 & 52 & \underline{65}
            & \textbf{74}\gain{22} \\
        Dump Bin Bigbin
            & 0 & 0 & 0 & 0
            & \textbf{37}\gain{37} \\
        Handover Block
            & 0 & 0 & 1 & \underline{17}
            & \textbf{20}\gain{19} \\
        Hanging Mug
            & \underline{9} & 0 & 5 & \underline{9}
            & \textbf{25}\gain{20} \\
        Place Object Basket
            & \underline{16} & 6 & 11 & 13
            & \textbf{23}\gain{12} \\
        Put Bottles Dustbin
            & 20 & \textbf{43} & 17 & 23
            & \underline{27}\gain{10} \\
        \addlinespace[4pt]
        \multicolumn{6}{l}{\emph{Dynamic Tracking}} \\
        Beat Block Hammer
            & 10 & 9 & 10 & \underline{15}
            & \textbf{23}\gain{13} \\
        Move Stapler Pad
            & \underline{1} & 0 & 0 & \underline{1}
            & \textbf{3}\gain{3} \\
        Place Bread Basket
            & 6 & 7 & 8 & \underline{12}
            & \textbf{23}\gain{15} \\
        Place Container Plate
            & 15 & 8 & 22 & \underline{26}
            & \textbf{44}\gain{22} \\
        Place Mouse Pad
            & 4 & 1 & \underline{5} & 2
            & \textbf{10}\gain{5} \\
        Place Shoe
            & 20 & 6 & \underline{25} & 16
            & \textbf{52}\gain{27} \\
        \midrule
        35-task Average
            & 8.17 & 10.86 & 9.63
            & \underline{17.20}
            & \textbf{20.74}\gain{11.11} \\
        \bottomrule
    \end{tabular}
\end{table}

\endgroup

\begin{table}[!t]
    \centering
    \caption{Success rates on the RMBench M(1) setting (\%).}
    \label{tab:rmbench-main}
    \footnotesize
    \setlength{\tabcolsep}{7.2pt}
    \renewcommand{\arraystretch}{1.09}
    \begin{tabular}{lcccccc}
        \toprule
        Policy & Obs. & Rearr. & Put B. & Swap B. & Swap T. & Avg. \\
        \midrule
        DP
            & 1 & 0 & 0 & 11 & 20 & 6.4 \\
        ACT
            & 1 & 29 & 0 & 2 & 2 & 6.8 \\
        X-VLA
            & 9 & 13 & 18 & 16 & 3 & 11.8 \\
        MemER
            & 7 & 17 & 0 & 14 & 7 & 9.0 \\
        MemoryVLA
            & 0 & 22 & 50 & 17 & 9 & 19.6 \\
        Mem-0
            & 4 & 89 & 90 & 67 & 14 & 52.8 \\
        ChainVLA
            & 11 & \underline{93} & \underline{96} & 74 & 40 & 62.8 \\
        MaP-WAM
            & \underline{19} & 66 & \textbf{100} & \underline{97}
            & \underline{96} & \underline{75.6} \\
        \midrule
        \rowcolor{baselinebg}
        \(\pi_{0.5}\) (Baseline)
            & 9 & 13 & 11 & 24 & 15 & 14.4 \\
        \rowcolor{oursbg}
        PREM (Ours)
            & \textbf{22}\gain{13}
            & \textbf{100}\gain{87}
            & \textbf{100}\gain{89}
            & \textbf{98}\gain{74}
            & \textbf{99}\gain{84}
            & \textbf{83.8}\gain{69.4} \\
        \bottomrule
    \end{tabular}
\end{table}

\paragraph{Dynamic manipulation.}
\label{sec:dynamic-manipulation}

DOMINO evaluates manipulation under continuously evolving object
and interaction states. Table~\ref{tab:domino-main} reports an
average success rate of \(20.74\%\) for PREM. This exceeds
\(\pi_{0.5}\) by \(11.11\) percentage points and PUMA by \(3.54\)
percentage points. PREM improves 31 of the 35 tasks over
\(\pi_{0.5}\). Examples include Adjust Bottle
(\(52\%\rightarrow74\%\)) and Dump Bin Bigbin
(\(0\%\rightarrow37\%\)). These results show improved performance
on dynamic manipulation. Complete results for all 35 tasks are
reported in the appendix.

\paragraph{Memory-dependent manipulation.}
\label{sec:long-horizon-dependency}

RMBench M(1) requires earlier task-relevant visual evidence.
This evidence may no longer be visible in current observations.
Table~\ref{tab:rmbench-main} reports an average success rate of
\(83.8\%\) for PREM. This exceeds \(\pi_{0.5}\) and MaP-WAM by
\(69.4\) and \(8.2\) percentage points, respectively. PREM achieves
\(100\%\), \(100\%\), \(98\%\), and \(99\%\) on Rearrange Blocks,
Put Back Block, Swap Blocks, and Swap-T. These results support the
effectiveness of the complete PREM framework on memory-dependent
manipulation.

\paragraph{Inference efficiency.}
\label{sec:inference-overhead}

Table~\ref{tab:latency} reports a mean end-to-end inference latency
of \(51.5\) ms for PREM and \(43.8\) ms for \(\pi_{0.5}\).
PREM therefore incurs a mean inference overhead of \(7.7\) ms on
a single NVIDIA H800 GPU. This measurement includes visual-state
updating, prediction verification, and action-conditioned future
prediction.

\subsection{Ablation Studies}
\label{sec:ablation-studies}

\paragraph{Component ablations.}
\label{sec:component-ablations}

We conduct an incremental component ablation on DOMINO.
Table~\ref{tab:component-ablation} examines the RVM architecture,
predictive supervision for memory learning, and prediction-regulated
visual-state updating.

\emph{Future Prediction} denotes training with the action-conditioned
future-prediction objective. It provides temporal supervision for
memory learning. The predictor uses the incoming visual state,
visual features, and the action expert's latent action representation.
\emph{Regulated Update} denotes the use of normalized prediction
discrepancy to regulate visual-state updating.

Adding the RVM architecture raises success rate from \(9.63\%\) to
\(15.43\%\), a gain of \(5.80\) percentage points. Predictive
supervision then raises success rate to \(18.31\%\), a gain of
\(2.88\) percentage points. With this supervision retained,
prediction-regulated visual-state updating raises success rate to
\(20.74\%\), a gain of \(2.43\) percentage points. Both predictive
supervision for memory learning and prediction-regulated visual-state
updating improve performance in this incremental configuration.

\begin{table}[H]
    \centering
    \begin{minipage}[t]{0.39\textwidth}
        \centering
        \caption{End-to-end mean inference latency on one NVIDIA H800
        80GB GPU with batch size \(1\).}
        \label{tab:latency}
        \small
        \setlength{\tabcolsep}{6.5pt}
        \renewcommand{\arraystretch}{1.08}
        \begin{tabular}{lc}
            \toprule
            Model & Mean latency (ms) \\
            \midrule
            \(\pi_{0.5}\) & 43.8 \\
            \rowcolor{oursbg}
            PREM (Ours) & 51.5 \\
            PUMA & 57.4 \\
            Mem-0 & 96.6 \\
            MemoryVLA & 133.8 \\
            \bottomrule
        \end{tabular}
    \end{minipage}
    \hfill
    \begin{minipage}[t]{0.55\textwidth}
        \centering
        \caption{Component ablation on DOMINO.
        Future Pred. denotes predictive supervision for memory.
        Regulated Update denotes feedback for
        visual-state updating.}
        \label{tab:component-ablation}
        \small
        \setlength{\tabcolsep}{2.3pt}
        \renewcommand{\arraystretch}{1.08}
        \begin{tabular}{lcccc}
            \toprule
            Variant & RVM & \shortstack{Future\\Pred.}
            & \shortstack{Regulated\\Update}
            & \shortstack{DOMINO\\SR (\%)} \\
            \midrule
            \(\pi_{0.5}\) &  &  &  & 9.63 \\
            \(+\) RVM & \(\checkmark\) &  &  & 15.43 \\
            \(+\) Future Prediction
                & \(\checkmark\) & \(\checkmark\) &  & 18.31 \\
            \rowcolor{oursbg}
            PREM
                & \(\checkmark\) & \(\checkmark\) & \(\checkmark\)
                & \textbf{20.74} \\
            \bottomrule
        \end{tabular}
    \end{minipage}
\end{table}

\begin{figure}[!t]
    \centering
    \includegraphics[width=0.90\linewidth]
    {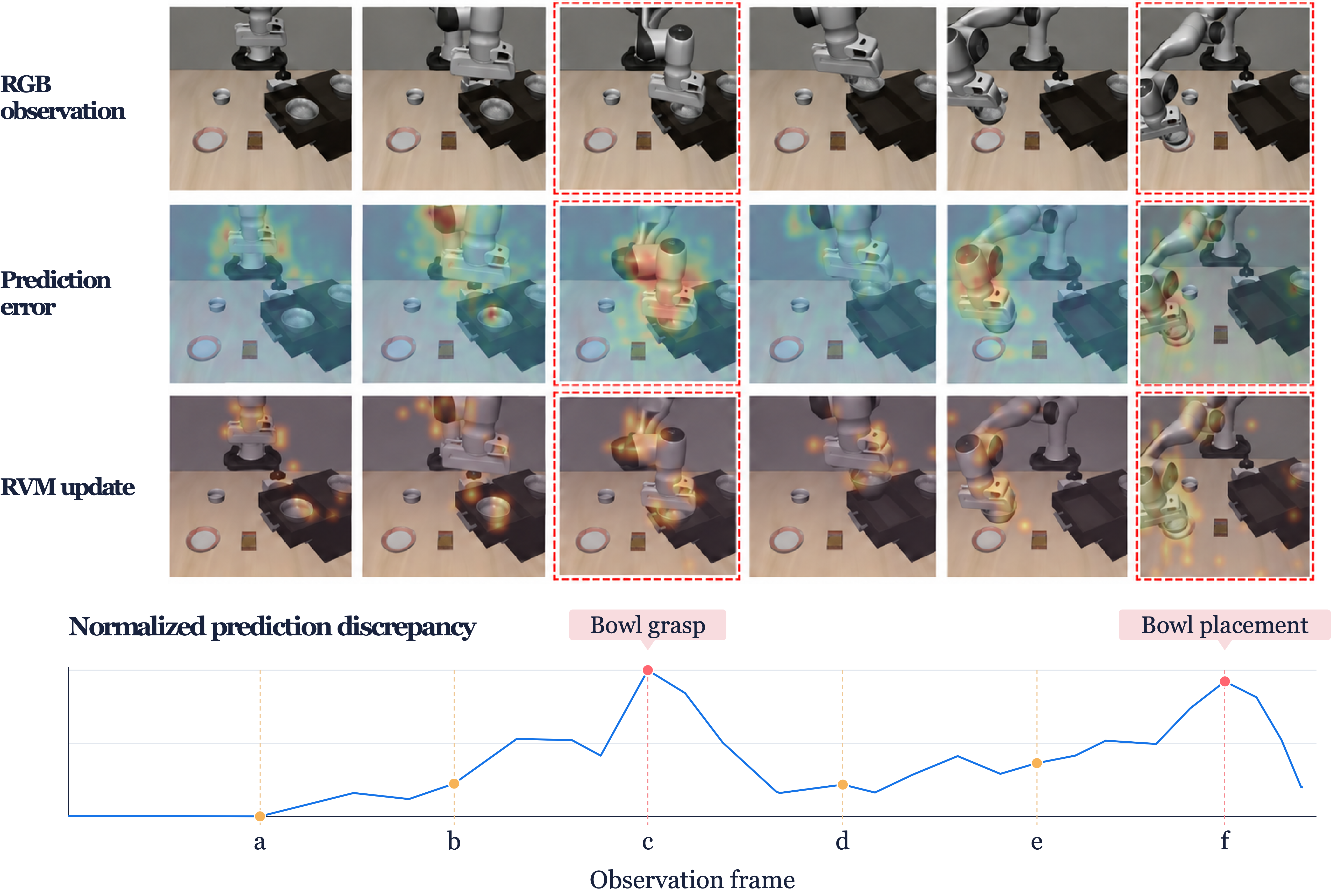}
    \caption{\textbf{Prediction discrepancy and visual-state updating.}
    A temporal prediction-discrepancy curve and visual-state update
    maps are shown for a successful LIBERO rollout. Spatial responses
    concentrate around the gripper and manipulated object near
    bowl grasp (\textbf{c}) and bowl placement (\textbf{f}).}
    \label{fig:mechanism-analysis}
\end{figure}

\paragraph{Mechanism analysis.}
\label{sec:mechanism-analysis}

\emph{Prediction discrepancy and visual-state updating.}
Figure~\ref{fig:mechanism-analysis} visualizes a successful LIBERO
rollout. The curve summarizes the temporal variation of prediction discrepancy over the interaction.. Larger values indicate
moments with more substantial visual changes, such as grasping, object
motion, occlusion changes, or placement, while smaller values indicate
more stable interaction. The orange markers and vertical lines
correspond to the six observations shown above. The spatial responses
concentrate around the gripper and manipulated object near bowl grasp
(\textbf{c}) and bowl placement (\textbf{f}), consistent with using
prediction discrepancy to guide visual-state updating.

\section{Conclusion}
\label{sec:conclusion}

We presented PREM, a framework that uses action-conditioned future
prediction to guide both memory learning and visual-state updating
for robot manipulation. Predictive supervision encourages the visual
state to retain historical information useful for predicting
subsequent visual features. Prediction discrepancy provides spatial
feedback for visual-state updating. The updated visual state supplies
temporal context for action generation alongside current observations.
Built on $\pi_{0.5}$, PREM achieves average success rates of $20.74\%$
on DOMINO and $83.8\%$ on RMBench M(1). It also improves general
manipulation on LIBERO and robustness to spatial perturbations on
LIBERO-Pro. PREM incurs a mean inference overhead of $7.7\,\mathrm{ms}$
on a single NVIDIA H800 GPU. These results support the joint use of
predictive supervision and prediction-discrepancy feedback for temporal
modeling in VLA policies. Together, these results show that future
prediction can both supervise visual-memory learning and regulate
visual-state updating in recurrent VLA policies. PREM requires no
keyframe or event annotations; its update signal comes directly from
future observations.

\clearpage
\subsection*{AI use statement}
In this work, generative AI tools were used to edit human-written text
for clarity and readability, assist with LaTeX typesetting and page
layout, and adjust the placement and sizing of existing figures and
tables. They were not used to generate or modify experimental data or
scientific-figure content, generate synthetic data, formulate the
research hypotheses or mathematical claims, develop the theoretical or
conceptual framework, design the methodology or experiments, implement
the proposed method, clean or reformat datasets, perform qualitative
data analysis, or interpret the experimental results. The authors
manually reviewed and verified all AI-assisted text, figures, numerical
values, citations, and formatting against the underlying implementation
and experimental records. The authors take full responsibility for the
final content of the paper and all artifacts produced with the assistance
of generative AI.

\subsection*{Ethics Statement}

This work follows the principles of the ICLR Code of Ethics.
We aim to conduct and report our research with scientific integrity,
transparency, and reproducibility. Experimental results are reported
without fabrication, falsification, or intentional misrepresentation,
and sufficient implementation and evaluation details are provided to
facilitate verification and reproduction. We appropriately acknowledge
prior work and use existing benchmarks, datasets, and associated
resources in accordance with their intended research purposes and
applicable licenses. This study does not involve human participants
or the collection of personal information.

\subsection*{Reproducibility statement}
We support reproducibility by documenting the model architecture,
prediction-regulated update rule, future-prediction objective, and
training formulation in Section~\ref{sec:method}. Section~\ref{sec:experiments}
specifies the benchmarks, baselines, training-data sizes, optimization
schedule, evaluation metrics, and latency protocol. The appendix provides
implementation-level architectural dimensions and hyperparameters
(Appendix~\ref{app:implementation-details}), benchmark-specific training
and evaluation procedures (Appendices~\ref{app:libero-protocol}--
\ref{app:rmbench-protocol}), evaluation seeds where applicable, and complete
per-task DOMINO results (Appendix~\ref{app:complete-results}). Code and
configuration files will be included in the supplementary material where
permitted to enable independent reproduction of the reported experiments.

\bibliography{references}
\bibliographystyle{iclr2027_conference}

\appendix
\section{Additional Related Work}
\label{sec:related-work}

\subsection{Temporal Memory for Robotic Manipulation}
\label{sec:related-temporal-modeling}

Generalist robot learning provides the broader setting for temporal modeling.
RT-1 studies scalable robot control from diverse demonstrations, while RT-2
transfers vision--language pretraining to action generation
~\citep{brohan2022rt1,brohan2023rt2}. PaLM-E integrates continuous sensory
inputs with language models for embodied reasoning and planning
~\citep{driess2023palme}. Open X-Embodiment aggregates robot data across
platforms~\citep{oneill2023openx}, while Octo and LingBot-VLA develop generalist
policies using diverse robot datasets~\citep{octo2024,wu2026lingbot}.
These advances provide a foundation for general-purpose manipulation; our
focus is how visual evidence is retained and updated during interaction.

Temporal manipulation policies differ primarily in how they retain history.
Short-window methods repeatedly encode recent observations: RoboFlamingo and
CronusVLA aggregate multiple frames, while DynamicVLA, PUMA, and Static In,
Dynamic Out target moving-object manipulation through continuous inference,
optical-flow-conditioned prediction, or future-pose estimation
~\citep{li2023roboflamingo,li2025cronusvla,xie2026dynamicvla,
fang2026domino,shin2026sido}. These approaches provide useful local motion cues,
but the retained history remains bounded by the chosen context window.

Longer-range methods instead maintain explicit memory. MemoryVLA and SAM2Act+
use visual or perceptual-cognitive memory banks, MemoryVLA++ combines memory
with imagined future latents, and ChainVLA propagates an execution state while
retaining sparse event memory~\citep{shi2025memoryvla,fang2025sam2act,
shi2026memoryvlapp,huang2026chainvla}. EventVLA writes observations according to
a predicted keyframe schedule~\citep{yang2026eventvla}. Such designs can retain
evidence over long intervals, but require explicit storage, consolidation, or
retrieval policies.

PREM follows a third design: it compresses the observation stream online into
a fixed-size recurrent visual state. Fixed-size memories have previously been
used for long-video recognition and action prediction
~\citep{wu2022memvit,ryoo2023ttm}, and our recurrent core is motivated by
Recurrent Video Masked Autoencoders~\citep{zoran2025rvm}. PREM adapts this
spatial carry to closed-loop manipulation and couples it with
action-conditioned prediction and delayed prediction verification. Thus, its
memory size does not grow with episode length, and prediction discrepancy
regulates visual-state updating at each spatial location.

\subsection{Predictive Representations and World Models}
\label{sec:related-predictive-modeling}

Future prediction is commonly used either as a training objective or as
additional inference-time context.
GR-1 connects video prediction with robot control by combining large-scale
video generative pretraining with joint prediction of future images and
robot actions~\citep{wu2023gr1}.
PUMA, Fast-WAM, DynaMo, and VLA-JEPA use
future or dynamics prediction to improve learned representations
~\citep{fang2026domino,yuan2026fastwam,cui2024dynamo,sun2026vlajepa}.
DreamVLA, DreamZero, ImageWAM, and MemoryVLA++ instead expose imagined future
content to action generation in various forms
~\citep{zhang2025dreamvla,ye2026dreamzero,zhang2026imagewam,
shi2026memoryvlapp}. Related latent-prediction principles also underlie V-JEPA
and V-JEPA 2~\citep{bardes2024vjepa,assran2025vjepa2}.

PREM uses prediction differently. The predicted future is retained until the
corresponding observation arrives, then used to regulate recurrent memory
writing rather than supplied as forecast context to the action expert.
Cambrian-S similarly uses prediction error to identify surprising events in
long videos~\citep{yang2025cambrians}; PREM instantiates this idea in a
closed-loop manipulation policy through its
\emph{prediction--verification--update} cycle. Its contribution is therefore
the coupling of action-conditioned future prediction, delayed verification,
and persistent visual-state control.

\subsection{Benchmarks for Temporal Manipulation}
\label{sec:related-temporal-benchmarks}

Beyond robotic control benchmarks, SpatialScore evaluates the spatial
understanding of multimodal models through diverse question-answering
tasks~\citep{wu2025spatialscore}. Such evaluations provide a complementary
view of spatial perception and reasoning, but do not directly test whether
a robot can retain historical evidence for closed-loop action selection.

LIBERO and CALVIN evaluate diverse or compositional manipulation
~\citep{liu2023libero,mees2022calvin}, but long episodes alone do not guarantee
that success depends on temporal evidence. LIBERO-Pro and subsequent benchmark
audits show that standard performance can partly exploit familiar layouts,
instructions, or action sequences~\citep{zhou2025liberopro,
jiang2026benchmarkaudit}. We therefore use LIBERO primarily to measure general
manipulation and LIBERO-Pro to test spatial robustness.

DOMINO and RMBench provide more targeted temporal diagnostics. DOMINO evaluates
dynamic interception and tracking under continuously evolving object states
~\citep{fang2026domino}, whereas RMBench constructs tasks whose solutions
explicitly require historical observations~\citep{chen2026rmbench}. We use them
as complementary tests of dynamic perception and long-range memory,
respectively; RMBench is not treated as a dynamic-environment benchmark.

\section{Implementation and Evaluation Details}
\label{app:implementation-details}

The following details use the notation of the main text. The index $t$
counts replanning decisions, while observation strides and future-target
offsets are measured in environment steps.

\subsection{Architecture and State Update}
\label{app:prem-implementation}

\paragraph{Visual representation and recurrent core.}
The recurrent pathway uses the current base-view image, corresponding to
the head-camera view for the Aloha setups. The pretrained PaliGemma visual
encoder produces a $16\times16$ grid of 2048-dimensional patch features.
A trainable \texttt{SpatialProjector} maps these features to
$X_t\in\mathbb{R}^{256\times512}$, giving $N=256$ spatial tokens and
memory width $d=512$. The RVM has two recurrent layers, eight attention
heads, and an MLP expansion ratio of two, corresponding to a feed-forward
hidden width of 1024.

Following the transformer--GRU formulation, the recurrent core uses
update and reset gates to construct its candidate state:
\begin{equation}
\begin{aligned}
u_t &= \sigma(X_tW_u+M_{t-1}U_u), \\
r_t &= \sigma(X_tW_r+M_{t-1}U_r), \\
G_t &= F_{\mathrm{cand}}
    \bigl(X_t,r_t\odot\operatorname{LN}(M_{t-1})\bigr), \\
\widetilde M_t
    &= (1-u_t)\odot M_{t-1}
       +u_t\odot\operatorname{LN}(G_t).
\end{aligned}
\label{eq:app-recurrent-core}
\end{equation}
Here, the learned projections act on the feature dimension of each
spatial token, and $G_t$ denotes the candidate features. The final
memory $M_t$ is obtained by applying the prediction-derived coefficient
to $\widetilde M_t-M_{t-1}$, as in
Eq.~\eqref{eq:regulated_update}.

\paragraph{Recurrent-memory readout.}
The updated RVM state contributes 256 policy-prefix tokens:
\begin{equation}
P_t^{\mathrm{RVM}}
=
\sigma(\alpha_{\mathrm{mem}})M_tW_{\mathrm{mem}},
\label{eq:app-memory-readout}
\end{equation}
where $W_{\mathrm{mem}}$ projects from width 512 to the policy token
width. The scalar $\alpha_{\mathrm{mem}}$ is learned and initialized
to $-4$. This readout corresponds to
$\mathcal P_{\mathrm{mem}}(M_t)$ in the main text.

\subsection{Action Representation and Future Predictor}
\label{app:action-conditioned-predictor}

\paragraph{Action-feature extraction.}
We reuse the final-layer action-token hidden states of the 18-layer
\texttt{gemma\_300m} Action Expert, whose hidden width is 1024.
Specifically, these are the final \texttt{action\_horizon} tokens of
\texttt{suffix\_out}, before the action output projection. No additional
action-expert forward pass is performed.

During training, the hidden states come from the single forward pass
at the randomly sampled flow-matching time. They are not obtained
from a complete iterative denoising trajectory. During inference,
they come from the final ODE/flow-denoising iteration.
Let $H_{\mathrm{act}}$ denote the predicted action-chunk length and
$H_{\mathrm{cond}}$ the number of leading action tokens used for
future prediction. Their benchmark-specific values are given in
Table~\ref{tab:app-action-conditioning}.

\begin{table}[t]
    \centering
    \caption{Action-token selection for future prediction. $B$ denotes
    batch size. Only the first $H_{\mathrm{cond}}$ action tokens are used.}
    \label{tab:app-action-conditioning}
    \small
    \setlength{\tabcolsep}{5pt}
    \begin{tabular}{@{}lccc@{}}
        \toprule
        Benchmark & Raw hidden shape & $H_{\mathrm{act}}$
                  & $H_{\mathrm{cond}}$ \\
        \midrule
        LIBERO  & $B\times10\times1024$ & 10 & 5 \\
        DOMINO  & $B\times50\times1024$ & 50 & 16 \\
        RMBench & $B\times50\times1024$ & 50 & 50 \\
        \bottomrule
    \end{tabular}
\end{table}

\paragraph{Temporal encoding and pooling.}
The selected hidden states are detached and projected from 1024 to
512 dimensions. After adding sinusoidal temporal encodings, a
one-layer temporal Transformer processes them using eight attention
heads and a feed-forward hidden width of 1024. The sequence is then
partitioned into eight ordered temporal bins, with masked mean pooling
within each bin. The resulting representation has shape
$z_t^{\mathrm{act}}\in\mathbb{R}^{B\times8\times512}$.
Empty bins are marked invalid and excluded from attention. In LIBERO,
five valid action-conditioning tokens yield five valid pooled tokens
among the eight output positions, giving a valid fraction of $5/8$.
Action tokens are not matched one-to-one with visual patches.

\paragraph{Query-based spatial decoding.}
The predictor is a two-block Transformer decoder with 256 learned
spatial queries of width 512. It receives the incoming memory
$M_{t-1}$, the current-step context $X_t$ defined in the main text,
and $z_t^{\mathrm{act}}$. The conditioning streams are mapped to a
common width of 512 and concatenated along the token dimension.
Each decoder block performs eight-head cross-attention from the spatial
queries to this context, a $512\rightarrow1024\rightarrow512$ MLP,
and self-attention among the spatial queries. The output is
$\hat X_{t+1}\in\mathbb{R}^{B\times256\times512}$.

The explicit memory input is the incoming $M_{t-1}$, not the newly
updated $M_t$. The action representation is obtained after action
generation has incorporated the updated memory. The forecast is
cached and compared with $X_{t+1}$ at the next replanning decision.

\subsection{Prediction Discrepancy and Temporal Training}
\label{app:prediction-training}

\paragraph{Normalized feature discrepancy.}
For feature vectors $u,v\in\mathbb{R}^{d}$, let
$\bar u=\operatorname{Norm}_{\epsilon_c}(u)$ and
$\bar v=\operatorname{Norm}_{\epsilon_c}(v)$ denote the token-wise
normalized features. The discrepancy used for prediction and memory
modulation is
\begin{equation}
\delta(u,v)
=
1-\operatorname{cos}_{\epsilon_c}(\bar u,\bar v)
+
\frac{\lambda_{\mathrm{dist}}}{d}
\sum_{c=1}^{d}
\operatorname{SmoothL1}_{\kappa}(\bar u_c-\bar v_c),
\label{eq:feature-discrepancy}
\end{equation}
where $\operatorname{cos}_{\epsilon_c}$ denotes cosine similarity
computed with numerically stabilized L2 normalization of its operands.
We use $\epsilon_c=10^{-6}$ for token normalization and cosine L2
normalization. The Smooth-L1 weight is $\lambda_{\mathrm{dist}}=0.1$,
and its transition parameter is $\kappa=0.1$. The Smooth-L1 term is
averaged over feature dimensions.

\paragraph{Spatial normalization and update coefficients.}
We compute $e_{t,i}=\delta(\hat X_{t,i},X_{t,i})$ and detach the error
before computing the memory-update coefficient. Writing
$e^{\perp}_{t,i}=\operatorname{sg}(e_{t,i})$, we obtain
\begin{equation}
\begin{aligned}
\mu_t &= \operatorname{median}_{j}e^{\perp}_{t,j}, \\
\rho_t &= \operatorname{median}_{j}|e^{\perp}_{t,j}-\mu_t|, \\
\xi_{t,i} &= \frac{e^{\perp}_{t,i}-\mu_t}
                   {\max(\rho_t,\epsilon_m)}.
\end{aligned}
\label{eq:app-normalized-error}
\end{equation}
The MAD denominator floor is $\epsilon_m=10^{-3}$. The coefficient
is then computed as in Eq.~\eqref{eq:update_coefficient}, using
$\beta_{\min}=0.25$, $\tau=2.0$, and $\gamma=0.5$.
The discrepancy and update-coefficient parameters are fixed
hyperparameters, not learned parameters.

\paragraph{Prediction objective and supervision.}
The target is the base/head-camera representation at the next
replanning observation, obtained from the same visual encoder and
trainable $2048\rightarrow512$ spatial projector. The projected target
features are detached before evaluating the discrepancy. There is
neither a separate target encoder nor an EMA target encoder.

Let $v_t$ mark a valid adjacent prediction--target pair and
$q_{t,i}\sim\operatorname{Bernoulli}(0.75)$ select supervised patches.
The prediction objective is
\begin{equation}
\mathcal L_{\mathrm{pred}}
=
\frac{
\sum_{t,i}v_tq_{t,i}\,
\delta\!\left(\hat X_{t+1,i},\operatorname{sg}(X_{t+1,i})\right)
}{
\sum_{t,i}v_tq_{t,i}
},
\label{eq:app-prediction-loss}
\end{equation}
with batch indices omitted for readability. The 75\% ratio refers to
the random selection of target patches contributing to this loss,
not to masking predictor inputs. Unselected target patches do not
contribute to the prediction loss. All 256 output tokens are available
at inference. The complete objective is
$\mathcal L=\mathcal L_{\mathrm{act}}+0.1\mathcal L_{\mathrm{pred}}$.

\paragraph{Warm-up and linear ramp.}
At optimizer step $s$, the modulation strength is
\begin{equation}
a(s)
=
\operatorname{clip}\!\left(
\frac{s-S_{\mathrm{warmup}}}{S_{\mathrm{ramp}}},0,1
\right),
\label{eq:app-modulation-schedule}
\end{equation}
which gives the effective update coefficient
\begin{equation}
\beta^{\mathrm{eff}}_{t,i}
=
1-a(s)+a(s)\beta_{t,i}.
\label{eq:app-effective-coefficient}
\end{equation}
During warm-up, $a=0$ and the full recurrent proposal is accepted.
During the ramp, the coefficient interpolates linearly toward the
prediction-regulated value. At inference, $a=1$.
Table~\ref{tab:app-temporal-config} reports the benchmark-specific
settings. These are memory-modulation schedules, not learning-rate
warm-up schedules.

\begin{table}[t]
    \centering
    \caption{Memory-modulation and temporal-training settings. Warm-up
    and ramp lengths count optimizer steps; $K$ counts replanning
    observations. Stride is measured in environment steps.}
    \label{tab:app-temporal-config}
    \small
    \setlength{\tabcolsep}{5pt}
    \begin{tabular}{@{}lcccc@{}}
        \toprule
        Benchmark & Warm-up & Ramp & $K$ & Stride \\
        \midrule
        LIBERO  & 2,000  & 2,000  & 16 & 5 \\
        DOMINO  & 10,000 & 10,000 & 16 & 16 \\
        RMBench & 1,000  & 2,000  & task-specific & 50 \\
        \bottomrule
    \end{tabular}
\end{table}

\paragraph{Causal unrolling and reset.}
Training uses \texttt{jax.lax.scan} to process each observation clip
causally. The recurrent carry is not detached within a clip, so
gradients propagate through the full unroll. Recurrent state is not
carried between training clips, and no burn-in prefix is used to
reconstruct pre-clip memory. The detached coefficient path described
above is separate from truncation of recurrent gradients.

The RVM state and forecast cache are reset to zero/invalid states at
clip and episode boundaries. The first valid observation initializes
$M_0=X_0$. Since there is no valid previous forecast, prediction-based
modulation is skipped at this observation. The first forecast uses the
reset incoming memory $M_{-1}$, which is excluded by its validity mask;
the current-step context and action representation remain available.

\subsection{LIBERO and LIBERO-Pro Protocol}
\label{app:libero-protocol}

\paragraph{Training configuration.}
We jointly train on the four LIBERO suites using 1,693 demonstrations,
a global batch size of 128, and 30,000 optimization steps. The learning
rate is cosine-annealed from $2\times10^{-5}$ to $1\times10^{-5}$.

\paragraph{Temporal sampling.}
Each clip contains $K=16$ replanning observations with a stride of five
environment steps. For a clip beginning at environment step $r_0$,
the observation indices are $[r_0,r_0+5,\ldots,r_0+75]$. Each future
target is offset by five environment steps. Clips may start within
an episode and initialize their RVM state from the first valid clip
observation, without burn-in. The predicted action horizon is ten,
and the first five action tokens condition the future predictor.
Memory-modulation warm-up and ramp lengths are 2,000 steps each.

\paragraph{LIBERO evaluation horizons.}
We use maximum episode horizons of 220, 280, 300, and 520 environment
steps for LIBERO-Spatial, LIBERO-Object, LIBERO-Goal, and LIBERO-10,
respectively. The corresponding LIBERO-Pro \texttt{\_swap} suites use
the same horizons. Both the standard and position-perturbed evaluations
contain ten tasks per suite and 50 episodes per task, yielding 2,000
episodes per evaluation setting.

\subsection{DOMINO Protocol}
\label{app:domino-protocol}

\paragraph{Training configuration and inputs.}
We initialize from the official $\pi_{0.5}$ base checkpoint and perform
full fine-tuning on the clean Aloha-AgileX Level-1 data from all 35 DOMINO
tasks, with at least 50 demonstrations per task. Training runs for
100,000 optimizer steps using eight-way FSDP, a global batch size of
64, and no gradient accumulation. Current observations contain head,
left-wrist, and right-wrist views; a missing right-wrist view is
represented by a mask. All views are aspect-ratio-preserving padded
to $224\times224$. The action horizon is 50.

\paragraph{Temporal sampling.}
Each clip contains 16 replanning observations at a stride of 16 raw
environment steps, $[r_0,r_0+16,\ldots,r_0+240]$. The future target
is the next replanning observation, offset by 16 steps. Clips may
start within an episode, with reset recurrent memory and no burn-in.
The first 16 action tokens condition future prediction.
Memory-modulation warm-up and ramp lengths are 10,000 steps each.

\paragraph{Evaluation protocol.}
Evaluation covers all 35 clean Level-1 tasks at DOMINO@$0.1$, with
100 accepted episodes per task. Task-specific horizons range from
400 to 1,700 environment steps. We use the exact training normalization
statistics, \texttt{adapt\_to\_pi=True}, a replanning interval of 16,
and \texttt{action\_speed\_scale=1.0}. Policy-server and client seeds
are both zero; simulator seeds begin at 100000 and are accepted
sequentially without duplication until each task reaches 100 valid
episodes. Each task is evaluated with a fresh policy server, with one
task assigned per GPU. The audit verifies 3,500 episodes, 3,500 videos,
unique accepted seeds, and the evaluation configuration.

\subsection{RMBench Protocol}
\label{app:rmbench-protocol}

\paragraph{Training configuration and inputs.}
We evaluate the five RMBench task families under the M(1)
memory-dependent setting. Each task is trained independently from
50 demonstrations for 20,000 optimization steps with a global batch
size of 32. Training uses eight GPUs with four-way FSDP and two-way
data parallelism, with gradient accumulation set to one.
The learning-rate warm-up lasts 2,000 steps; the learning rate is
cosine-annealed from $2\times10^{-5}$ to $1\times10^{-5}$.
The memory-modulation schedule instead uses 1,000 warm-up steps and
2,000 ramp steps, becoming fully enabled at optimizer step 3,000.
Current observations contain head, left-wrist, and right-wrist views.

\paragraph{Temporal sampling.}
All tasks use an action horizon, replanning interval, observation
stride, future-target offset, and action-conditioning horizon of
50 environment steps. The training runner uses strict temporal anchors,
with clips starting at episode step zero.

\paragraph{Gradient and padding handling.}
The recurrent carry is differentiated through the full $K$-step clip
without an internal detach. No recurrent state is transferred between
clips, and no burn-in is used. Clips shorter than the configured
length are padded, with padded observation positions marked by
\texttt{memory\_sequence\_valid=False}. For a terminal action chunk
with fewer than 50 valid actions, \texttt{action\_valid\_mask} excludes
repeated padding actions from the action loss.

\paragraph{Evaluation protocol.}
Evaluation uses 100 episodes per task. Recurrent states and pending
predictions are reset at every episode boundary.

\clearpage
\section{Complete Quantitative Results}
\label{app:complete-results}

The main paper reports representative DOMINO tasks. Table~\ref{tab:domino-full}
provides the complete 35-task comparison used to compute the aggregate result.

\subsection{DOMINO Per-task Results}
\label{app:domino-complete-results}

\makeatletter
\setlength{\@fptop}{0pt}
\makeatother
\begin{table}[H]
    \centering
    \caption{Complete per-task success rates (\%) on DOMINO Level-1.}
    \label{tab:domino-full}
    \footnotesize
    \begin{tabular}{l>{\columncolor{baselinebg}}c c>{\columncolor{oursbg}}c}
        \toprule
        Task & $\pi_{0.5}$ (Baseline) & PUMA & PREM (Ours) \\
        \midrule
        \multicolumn{4}{l}{\emph{Dynamic Interception}} \\
        Adjust Bottle            & 52 & 65 & \textbf{74}\gain{22} \\
        Dump Bin Bigbin          & 0  & 0  & \textbf{37}\gain{37} \\
        Grab Roller              & 10 & \textbf{33} & 26\gain{16} \\
        Handover Block           & 1  & \underline{17} & \textbf{20}\gain{19} \\
        Handover Mic             & 5  & \textbf{35} & \underline{26}\gain{21} \\
        Hanging Mug              & 5  & 9 & \textbf{25}\gain{20} \\
        Move Can Pot             & 7  & 22 & \textbf{29}\gain{22} \\
        Move Playing Card Away   & \textbf{8} & 6 & 7\dropresult{1} \\
        Place A2B Left           & 2  & \textbf{13} & 7\gain{5} \\
        Place A2B Right          & 1  & \textbf{8} & 4\gain{3} \\
        Place Bread Skillet      & 9  & \textbf{19} & 9\gain{0} \\
        Place Can Basket         & 6  & \textbf{14} & 13\gain{7} \\
        Place Object Basket      & 11 & 13 & \textbf{23}\gain{12} \\
        Put Bottles Dustbin      & 17 & 23 & \textbf{27}\gain{10} \\
        Put Object Cabinet       & 11 & \textbf{34} & 29\gain{18} \\
        Rotate QRcode            & 4  & \textbf{14} & 4\gain{0} \\
        Scan Object              & 1  & 5 & \textbf{6}\gain{5} \\
        Shake Bottle             & 33 & 55 & \textbf{55}\gain{22} \\
        Shake Bottle Horiz.      & 42 & \textbf{75} & 60\gain{18} \\
        \multicolumn{4}{l}{\emph{Dynamic Tracking}} \\
        Beat Block Hammer        & 10 & 15 & \textbf{23}\gain{13} \\
        Click Alarmclock         & \textbf{14} & 4 & 6\dropresult{8} \\
        Click Bell               & 0 & \underline{3} & \textbf{4}\gain{4} \\
        Move Pillbottle Pad      & 6 & \underline{14} & \textbf{21}\gain{15} \\
        Move Stapler Pad         & 0 & 1 & \textbf{3}\gain{3} \\
        Place Bread Basket       & 8 & 12 & \textbf{23}\gain{15} \\
        Place Container Plate    & 22 & 26 & \textbf{44}\gain{22} \\
        Place Empty Cup          & 2 & \underline{7} & \textbf{10}\gain{8} \\
        Place Fan                & 2 & \textbf{8} & 4\gain{2} \\
        Place Mouse Pad          & 5 & 2 & \textbf{10}\gain{5} \\
        Place Object Scale       & 3 & 4 & \textbf{8}\gain{5} \\
        Place Object Stand       & \textbf{1} & \textbf{1} & \textbf{11}\gain{10} \\
        Place Phone Stand        & 2 & 6 & \textbf{10}\gain{8} \\
        Place Shoe               & 25 & 16 & \textbf{52}\gain{27} \\
        Press Stapler            & 6 & \textbf{10} & 7\gain{1} \\
        Stamp Seal               & 6 & \textbf{13} & 9\gain{3} \\
        35-task Average          & 9.63 & 17.20 & \textbf{20.74}\gain{11.11} \\
        \bottomrule
    \end{tabular}
\end{table}

\clearpage
\section{Qualitative Rollouts}
\label{app:qualitative-rollouts}
\raggedbottom

We provide successful rollout visualizations across LIBERO, DOMINO, and
RMBench to complement the quantitative results and illustrate the distinct
manipulation capabilities evaluated by each benchmark.

\subsection{LIBERO}
\label{app:libero-qualitative-rollouts}

We visualize representative successful PREM rollouts from the four standard
LIBERO suites. Frames progress from left to right and cover spatial transfer,
object-centric placement, goal-conditioned interaction, and longer-horizon
composition.

\begin{figure}[H]
    \centering
    \includegraphics[width=\linewidth]
    {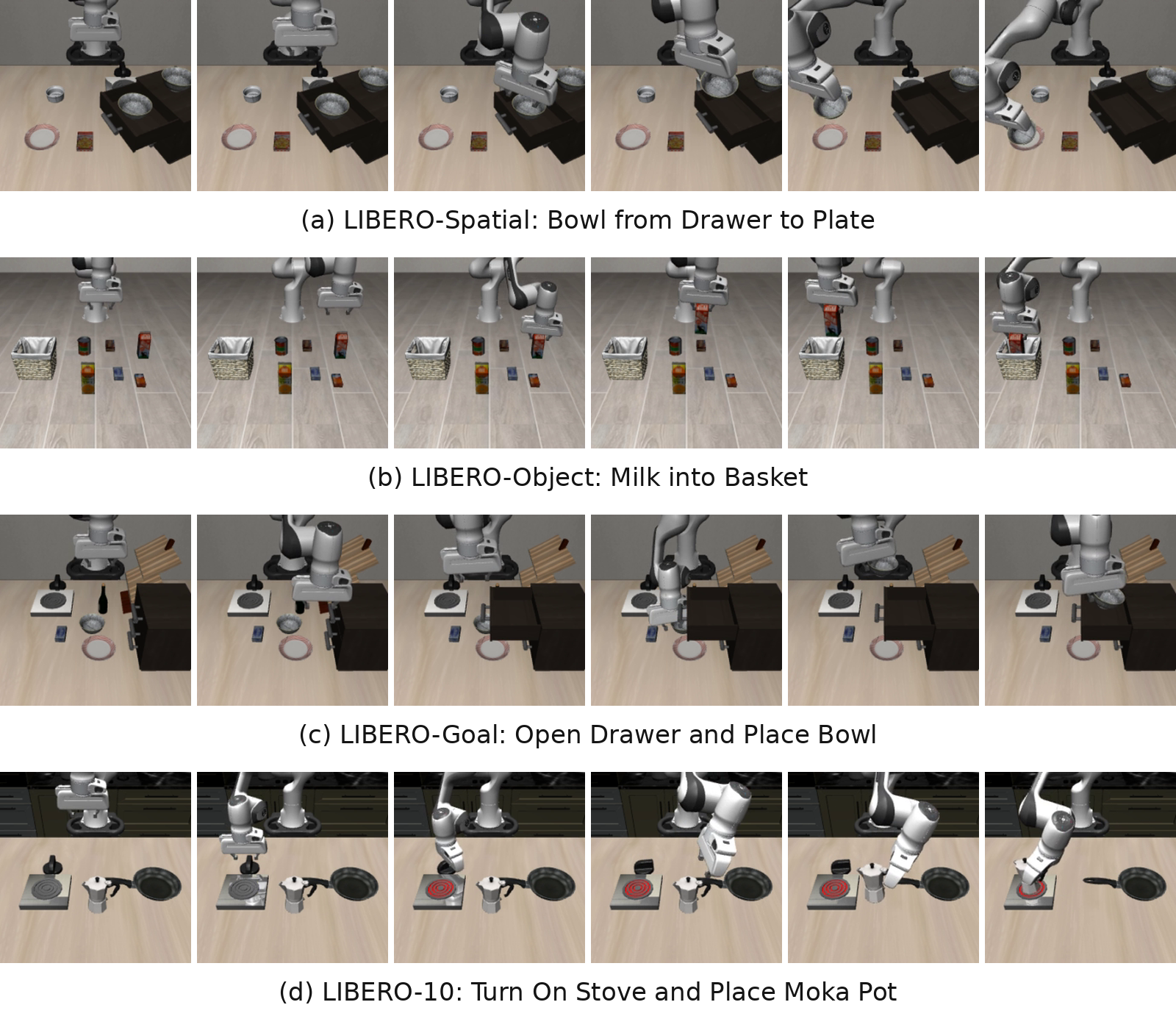}
    \caption{\textbf{Representative successful rollouts on LIBERO.}
    The rows show LIBERO-Spatial, LIBERO-Object, LIBERO-Goal, and LIBERO-10,
    respectively.}
    \label{fig:libero-successful-rollouts}
\end{figure}

\clearpage
\subsection{DOMINO}
\label{app:domino-qualitative-rollouts}

We additionally visualize successful PREM rollouts from the two official
DOMINO task categories. Frames progress from left to right, and the green
arrows indicate task-relevant object motion.

\begin{figure}[H]
    \centering
    \includegraphics[width=\linewidth]
    {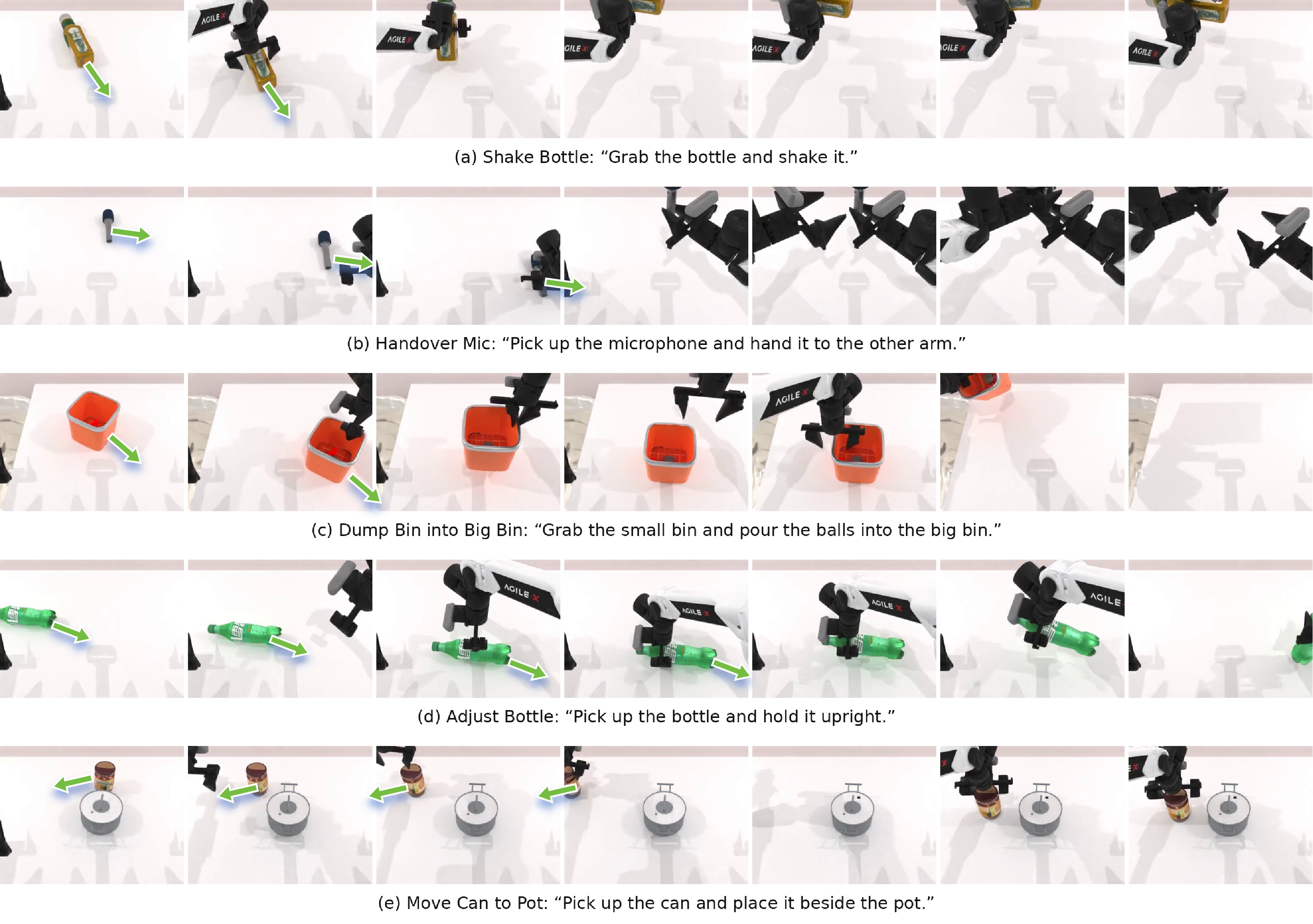}
    \caption{\textbf{Successful Dynamic Interception rollouts on DOMINO.}
    The five examples visualize Shake Bottle, Handover Mic, Dump Bin into Big
    Bin, Adjust Bottle, and Move Can to Pot. These tasks require the policy to
    react to objects or interaction states that evolve during execution.}
    \label{fig:domino-dynamic-interception-rollouts}
\end{figure}

\begin{figure}[H]
    \centering
    \includegraphics[width=\linewidth]
    {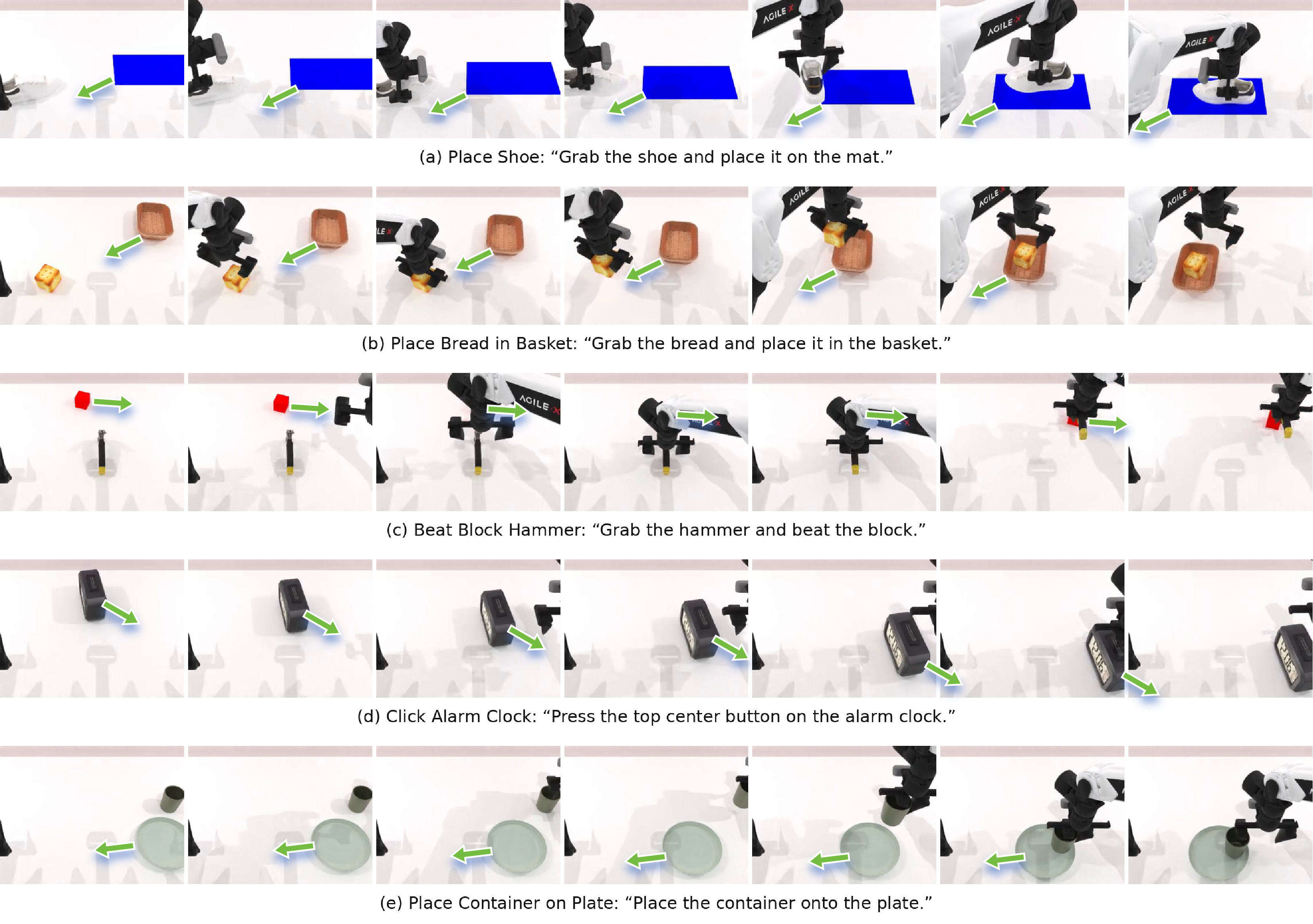}
    \caption{\textbf{Successful Dynamic Tracking rollouts on DOMINO.}
    The five examples visualize Place Shoe, Place Bread in Basket, Beat Block
    Hammer, Click Alarm Clock, and Place Container on Plate. Successful
    execution requires continuously tracking the task-relevant object as its
    spatial relationship with the robot changes.}
    \label{fig:domino-dynamic-tracking-rollouts}
\end{figure}

\clearpage
\subsection{RMBench}
\label{app:rmbench-qualitative-rollouts}

We visualize successful PREM rollouts for all five task families in the
RMBench M(1) setting. Frames progress from left to right and illustrate how the
policy retains task-relevant information across temporally separated
observations and actions.

The corresponding PREM success rates are (22\%), (100\%), (100\%),
(98\%), and (99\%) for Observe and Pick Up, Rearrange Blocks, Put Back
Block, Swap Blocks, and Swap-T, respectively, yielding an (83.8\%)
macro-average.

\begin{figure}[H]
    \centering
    \includegraphics[width=\linewidth]
    {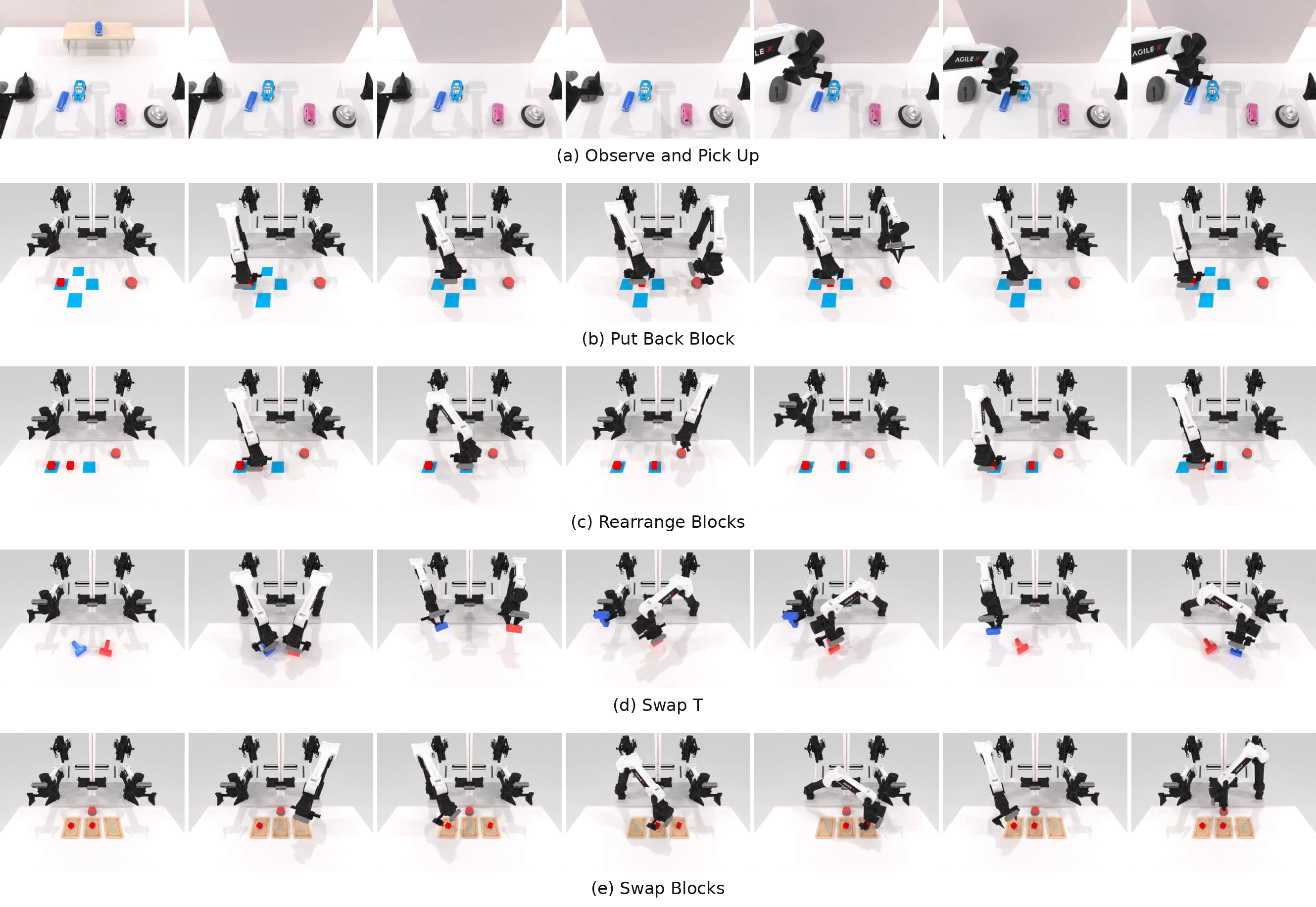}
    \caption{\textbf{Successful rollouts across the five RMBench M(1) task
    families.} The rows visualize Observe and Pick Up, Put Back Block,
    Rearrange Blocks, Swap-T, and Swap Blocks. These tasks require retaining
    object identities, initial spatial configurations, or earlier interaction
    states that cannot always be recovered from the current observation.}
    \label{fig:rmbench-successful-rollouts}
\end{figure}

\end{document}